\documentclass[journal]{IEEEtran}
\usepackage{cite}

\ifCLASSINFOpdf
  \usepackage[pdftex]{graphicx}
\else
\fi

\usepackage[export]{adjustbox}

\ifCLASSOPTIONcompsoc
  \usepackage[caption=false,font=normalsize,labelfont=sf,textfont=sf]{subfig}
\else
  \usepackage[caption=false,font=footnotesize]{subfig}
\fi

\usepackage[export]{adjustbox}
\usepackage{multirow}
\usepackage{array}
\usepackage{amsmath}
\usepackage[T1]{fontenc}
\usepackage[utf8]{inputenc}
\usepackage[english]{babel}
\usepackage[table]{xcolor}
\usepackage{collcell}
\usepackage{hhline}
\usepackage{pgf}

\def\colorModel{hsb} 

\newcommand\ColCell[1]{
	\pgfmathparse{#1<50?1:0}  
	\ifnum\pgfmathresult=0\relax\color{white}\fi
	\pgfmathsetmacro\compA{0}      
	\pgfmathsetmacro\compB{#1/100} 
	\pgfmathsetmacro\compC{1}   
	\edef\x{\noexpand\centering\noexpand\cellcolor[\colorModel]{\compA,\compB,\compC}}\x #1
} 
\newcolumntype{E}{>{\collectcell\ColCell}m{0.35cm}<{\endcollectcell}}  

\begin{document}
%
%
\title{Considerations for a PAP Smear Image Analysis System with CNN Features}


\author{Srishti~Gautam,
        Harinarayan~K. K.,
        Nirmal~Jith,
        Anil~K. Sao,
        Arnav~Bhavsar,
        and~Adarsh~Natarajan
\thanks{Srishti~Gautam (email: srishti\_gautam@students.iitmandi.ac.in), Anil~K. Sao and~Arnav~Bhavsar (email: anil/arnav@iitmandi.ac.in) are with Indian Institute of Technology, Mandi. }
\thanks{Harinarayan~K. K., Nirmal~Jith and~Adarsh~Natarajan are with Aindra Systems Pvt. Ltd., Bangalore.  (email: hari/nirmal/adarsh@iitmandi.ac.in)}}
\maketitle

\begin{abstract}
It has been shown that for automated PAP-smear image classification, nucleus features can be very informative. 
Therefore, the primary step for automated screening can be cell-nuclei detection followed by segmentation of nuclei in the resulting single cell PAP-smear images. We propose a patch based approach using CNN for segmentation of nuclei in single cell images. We then pose the question of ion of segmentation for classification using representation learning with CNN, and whether low-level CNN features may be useful for classification. We suggest a CNN-based feature level analysis and a transfer learning based approach for classification using both segmented as well full single cell images. We also propose a decision-tree based approach for classification. Experimental results demonstrate the effectiveness of the proposed algorithms individually (with low-level CNN features), and simultaneously proving the sufficiency of cell-nuclei detection (rather than  accurate segmentation) for classification. Thus, 
we propose a system for analysis of multi-cell PAP-smear images consisting of a simple nuclei detection algorithm followed by classification using transfer learning. 

\end{abstract}


%
\IEEEpeerreviewmaketitle

\section{Introduction}
%
%
%
%
\IEEEPARstart{C}{ervical} cancer continues to be one of the deadliest cancers among women worldwide, especially with it being the most common cause of death in developing countries \cite{chhabra,sreedevi2015,2014screening}. 
Every year, approximately 500,000 new cases are reported, out of which 85\% occur in developing countries, along with approximately 270,000 deaths worldwide \cite{chhabra}. The pre-cancerous lesions of cervical cancer take almost a decade to convert into the cancerous ones. Hence, despite the above facts, unlike many other cancers, it can be fully cured if detected early \cite{sreedevi2015}. 


For screening, traditional PAP-smear test continues to be prevalent, especially in the developing countries \cite{2014screening}. Due to vast differences in the morphology of cells (in size and regularity of nucleus) because of the cancerous/pre-cancerous changes, manual screening is reasonably straightforward. 
However, 
it has many drawbacks in terms of being tedious, time-consuming and expensive 
\cite{2014screening}. Also, there can be huge inter and intra observer variability \cite{pr}. Hence, automation is essential for development of a system with lower cost, adequate speed up and higher accuracy.


Automatic screening system using PAP-smear image analysis, traditionally comprises of three steps i.e cell (cytoplasm and nuclei) segmentation, feature extraction and cell classification. Segmentation seems important because the morphological changes associated with the degree of malignancy can be represented by features calculated from the segmented nuclei and cytoplasm, for example, the nucleus-cytoplasm ratio 
or texture features associated with chromatin pattern irregularity \cite{morph}. However, typically the segmentation of nucleus is more reliable than that of cytoplasm (possibly due to overlapping, occluded  cytoplasm in multi-cell images). Moreover, only the nucleus based features can also be extremely valuable and effective in cervical cancer screening \cite{icip}. Further, for some classification frameworks, a Region of Interest (ROI) / nuclei detection step in the images containing multiple cells may be used as a substitute to accurate segmentation step. These detected nuclei can be used for classification. 

While the segmentation process is more rigorous, but can assist the classifier to focus only on the features of the object in question, detection is relatively easier. Additionally, due to presence of background in the detected sub-images, some background contextual information is available when using detected cells as opposed to accurately segmented ones. Therefore, an interesting question to consider is that whether an accurate segmentation necessary for classification with contemporary frameworks (e.g. CNN)? Having said that, we acknowledge that segmentation can be useful on its own, in case one considers manual interventions, and for medical education and training applications. With regards to the CNN based classification, we note that unlike in standard computer vision applications, cell images, arguably do not contain high-level semantics. Thus, we also explore the effect of high-level CNN features vs. the low-level CNN features, where the latter can enable the system to be more efficient. 



Thus, noting the above aspects, for an overall system development, we propose algorithms for 1) Detection of nuclei in multi-cell images in single cell images, 2) Segmentation of nuclei in single cell images, 3) Classification strategies considering both accurate nuclei segmentation, and nuclei detection (involving some cellular background pixels), and also considering CNN features from different layers. 

\subsection{Related work: Segmentation}
\label{sec:seg_lit}
Numerous works have been reported for cervical cell nucleus segmentation, indicating its importance. Phoulady et. al \cite{icip,rw3} uses adaptive multilevel thresholding and ellipse fitting followed by iterative thresholding and subsequent binarizations but on a different problem of segmenting overlapping cells. Cheng et. al uses HSV color space and color clustering. 
Genctav et. al \cite{pr} uses multi-scale hierarchical segmentation algorithm. 
Ref \cite{11} uses patch-based fuzzy C-means (FCM) clustering technique, where, on the over-segmented image obtained from FCM, a threshold is applied to classify the FCM cluster centers as nucleus and background. 
A superpixel-based Markov random field (MRF) framework with a gap-search algorithm is proposed in \cite{MRF}. Bora et. al \cite{bora} uses Wavelet and Haar transform along with MSER (maximally stable extremal region). 
In recent years, deep learning techniques have also been explored in this area. 
 Multiscale CNNs are used in \cite{mscn} along with superpixels for multi-cellular images. Song et. al. also uses neural networks in \cite{dcnn} for nucleus and cytoplasm segmentation, and multiple scale deep CNNs \cite{song1,song2} for overlapping cell segmentation. However, most of the above approaches use non-public datasets, or public datasets lacking variation of normal and abnormal cell images (which typically have very different characteristics). For example, ISBI segmentation challenge dataset \cite{rw3} for overlapping cells, has no distinction between normal and abnormal slides. Only the approaches in \cite{11,MRF} use a publicly available dataset suitable for full cervical cancer diagnosis i.e. segmentation and classification of cells into normal vs abnormal. 

In this work, we report a CNN based segmentation approach, which works on selectively pre-processed cell images depending on the homogeneity of the nucleus, post which the pre-processed and non-pre-processed cells are segmented with two different CNNs. Considering low amount of variety in data, such a selective preprocessing helps each of the CNNs to learn the data characteristics better. 

\subsection{Related work: Classification}
\label{sec:class_lit}

Recently, deep learning and CNNs have gained the center stage for various classification problems \cite{vggnet, alexnet}. They have also gained some popularity in the medical imaging applications \cite{cnn_dl1, cnn_dl2, deepcell}. An important feature of CNN is a reduced dependency on an exact segmentation for classification. More specifically, an approximate segmentation (or ROI detection) can be considered to be sufficient for classification. This is especially important for the application of classification in medical imaging. However, there are a few drawbacks associated with training a CNN from scratch for a particular problem, for example, the availability of a very few number of annotated images especially in the case of medical datasets \cite{herlev}, requirement of a huge number of days/weeks to train etc. Transfer learning \cite{tl1} has proved to be very effective in overcoming these limitations, both in medical \cite{tl2, tl3} and non-medical domains \cite{tl1}. On account of the CNN features being more generic at early layers \cite{tl1} and having been already learned on a million images \cite{alexnet}, these can be used to train the subsequent layers of application-specific CNN. This reduces the chances of overfitting as well as the overall training time of CNN. Recently, both methodologies i.e training a CNN from scratch as well as transfer learning have also been applied on PAP-smear images \cite{icvgip, deeppap}. 

In this work, we consider deep learning methods for classification, using transfer learning on Alexnet \cite{alexnet}, on both segmented as well as non-segmented single cell images. Alexnet is selected considering the need for a smaller architecture, enabling an efficient processing in medical systems. We also propose a combination of decision-tree based classification with transfer learning. Finally, the transfer learning approach is applied on the detected cells from multi-cell images and is also shown to perform effectively.

\begin{figure*}[!t]
	{
		\centering
		\includegraphics[scale=0.5]{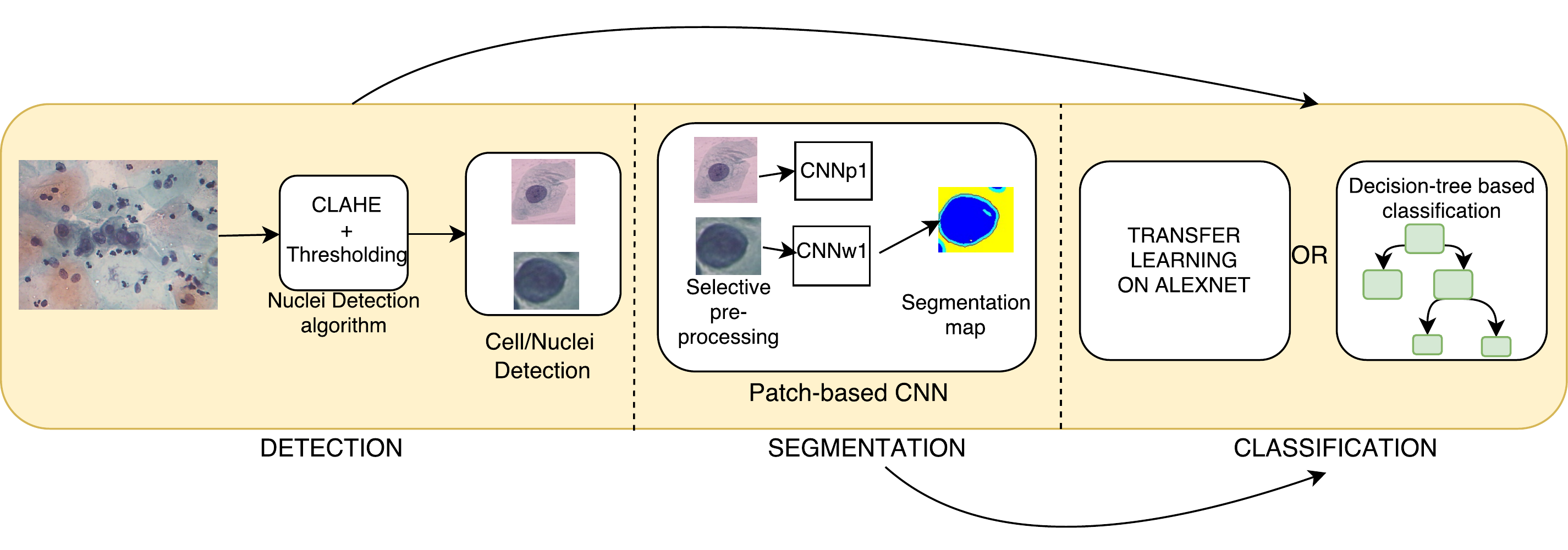}
		\caption{Proposed system for automated analysis of PAP-smear images.}
		\label{fig:block}
	}
\end{figure*}

Thus, the overall contributions of this work are: 1) We propose a patch-based CNN approach for segmentation using selective pre-processing and show that the proposed selectiveness is effective for nuclei segmentation in single-cell images. 2) We explore the classification results with transfer learning from the features extracted from different CNN layers in Alexnet \cite{alexnet}, and demonstrate that the low-level features can be more effective. 3) We demonstrate that the easier cell-nuclei detection can be more effective than an accurate segmentation for CNN-based classification. 4) We introduce a decision-tree based classification, which outperforms a simple multi-class classification with transfer learning. 5) We consider various classification scenarios and demonstrate state-of-the-art classification results. 

A preliminary version of this work has been reported \cite{myspie}.

\section{Dataset}
In this work, we have used two datasets for our experimentation, one for single cell images and another for multi-cell images. The latter is relatively more noisy and artifact-prone. 
\subsection{Herlev dataset}
The first dataset on which we have evaluated our algorithms is Herlev PAP-smear dataset \cite{herlev}. It consists of 917 cell images whose description in the increasing order of abnormality is given in Table \ref{tbl:herlev}. Each image in Herlev dataset consists of a single nuclei. It is a publicly available dataset collected at Herlev University Hospital by a digital camera and microscope. 
\begin{table*}[!t]
	\centering
	\caption{ Sample cervical cells in Herlev dataset
	}
	\label{tbl:herlev}
	\begin{tabular}{>{\centering\arraybackslash}p{1.5cm}|>{\centering\arraybackslash}p{1.5cm} >{\centering\arraybackslash}p{1.5cm} >{\centering\arraybackslash}p{1cm}|>{\centering\arraybackslash}p{1.5cm} >{\centering\arraybackslash}p{1.2cm} >{\centering\arraybackslash}p{1.2cm} >{\centering\arraybackslash}p{1.2cm}}
		\hline
		&\multicolumn{3}{c|}{\textbf{\scriptsize{Normal}}} & \multicolumn{4}{c}{\textbf{\scriptsize{Abnormal}}} \\ \hline 
	Classes&	\scriptsize{Superficial Squamous (nsup) } & \scriptsize{Intermediate Squamous (nint)} & \scriptsize{Columnar (ncol)} & \scriptsize{Light dysplasia (ldys)} & \scriptsize{Moderate dysplasia (mdys)} & \scriptsize{Severe dysplasia (sdys)} & \scriptsize{Carcinoma in situ (cis)} \\ \hline
				{{Sample cells}} &
		\begin{minipage}{1.5cm}
			\centering\includegraphics[scale=0.2]{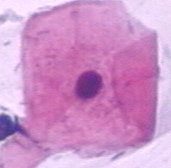}
		\end{minipage}
		&
		\begin{minipage}{1.5cm}
			\centering\includegraphics[scale=0.2]{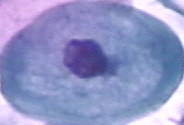}
		\end{minipage}
		& 
		\begin{minipage}{1cm}
			\centering\includegraphics[scale=0.2]{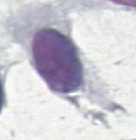}
		\end{minipage}
		&
		\begin{minipage}{1.5cm}
			\centering\includegraphics[scale=0.2]{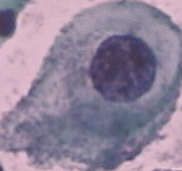}
		\end{minipage}
		&
		\begin{minipage}{1.2cm}
			\centering\includegraphics[scale=0.2]{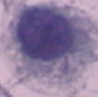}
		\end{minipage}
		&
		\begin{minipage}{1.2cm}
			\centering\includegraphics[scale=0.2]{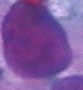}
		\end{minipage}
		& 
		\begin{minipage}{1.2cm}
			\centering\includegraphics[scale=0.2]{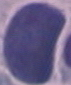}
		\end{minipage}
		\\ 
		\hline
		Total cells & \scriptsize74 & \scriptsize70 & \centering\scriptsize98 & \scriptsize182 & \scriptsize146 & \scriptsize197 & \scriptsize150 \\  \hline
	\end{tabular}
\end{table*}

\subsection{Aindra dataset}
80 multi-cell images were collected by Aindra Systems Pvt. ltd., Bangalore, India, from an oncology center. Staining and preparation of slides was done at the same center. 
The images are labeled into 4 classes i.e Normal, Low-grade squamous intraepithelial lesion (LSIL), High-grade squamous intraepithelial lesion (HSIL) and Squamous cell carcinoma (SCC). The nuclei in these images have been annotated by doctors. The sample images in the increasing order of abnormality are shown in Table \ref{tbl:Aindra}.
\begin{table*}[!t]
	\centering
	\caption{ Sample cervical cells in Aindra dataset. 
	}
	\label{tbl:Aindra}
	\begin{tabular}{>{\centering\arraybackslash}p{1.6cm}|>{\centering\arraybackslash}p{2.7cm}|>{\centering\arraybackslash}p{2.7cm} >{\centering\arraybackslash}p{2.7cm} >{\centering\arraybackslash}p{2.7cm}}
		\hline
	&	{\textbf{\scriptsize{Normal}}} & \multicolumn{3}{c}{\textbf{\scriptsize{Abnormal}}} \\ \hline 
	Classes&	& \scriptsize{LSIL} & \scriptsize{HSIL} & \scriptsize{SCC} \\ 
	Sample cells&	\begin{minipage}{2.7cm}
			\centering\includegraphics[scale=0.10]{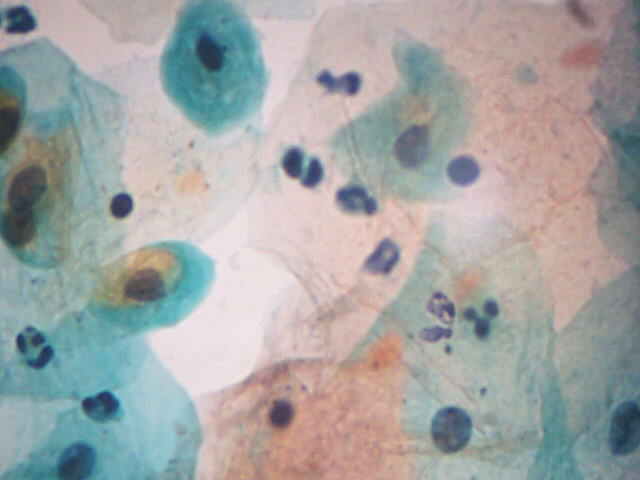}
		\end{minipage}
		&
		\begin{minipage}{2.7cm}
			\centering\includegraphics[scale=0.10]{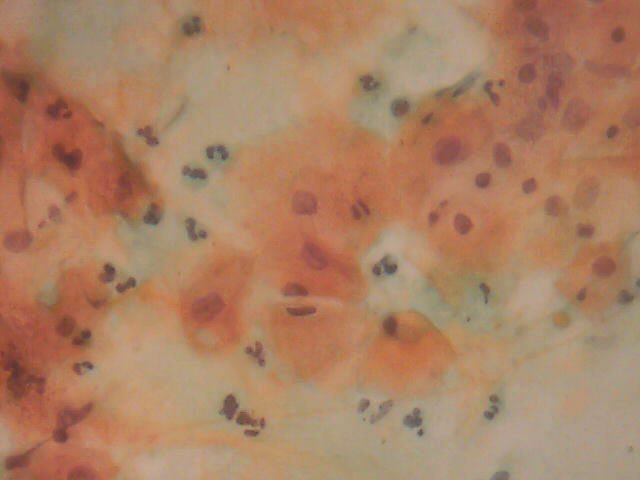}
		\end{minipage}
		& 
		\begin{minipage}{2.7cm}
			\centering\includegraphics[scale=0.10]{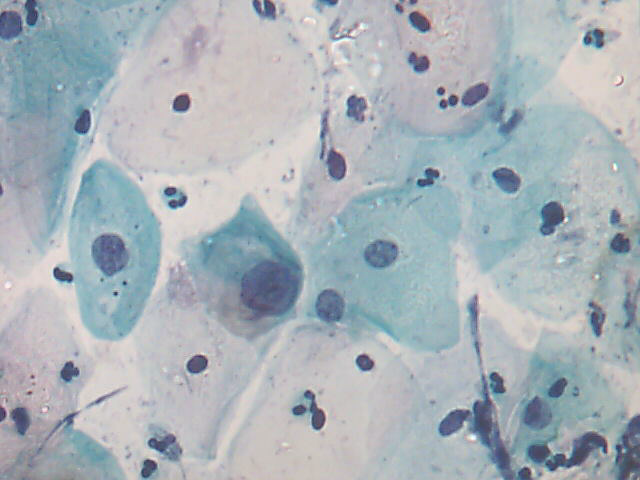}
		\end{minipage}
		&
		\begin{minipage}{2.7cm}
			\centering\includegraphics[scale=0.10]{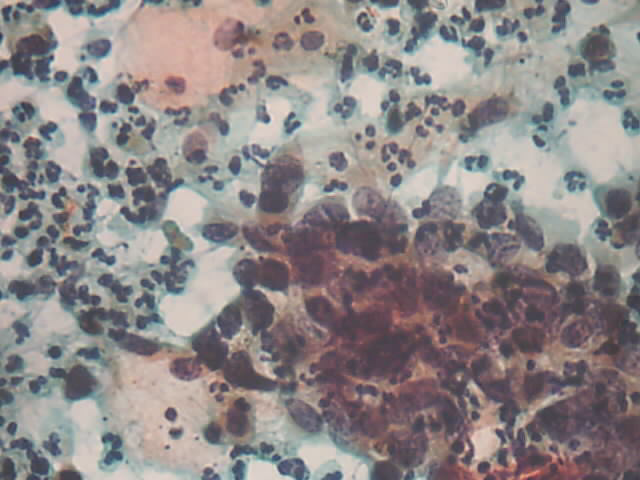}
		\end{minipage}
		\\ 
		\hline
Total cells &	\scriptsize36 & \scriptsize13 & \scriptsize21 & \scriptsize10 \\ \hline
	\end{tabular}
\end{table*}

\section{Proposed methodology}
\label{sec:method}
We now describe the proposed system which consists of three methods i.e detection of nuclei, segmentation of nuclei, and classification of segmented/detected nuclei via deep-learning approaches. 
The block diagram of the proposed system is given in Figure \ref{fig:block}. 

\subsection{Detection of nuclei in multi-cell images}
\begin{figure}[!h]
	\begin{center}
		\subfloat[]{\includegraphics[scale=0.13]{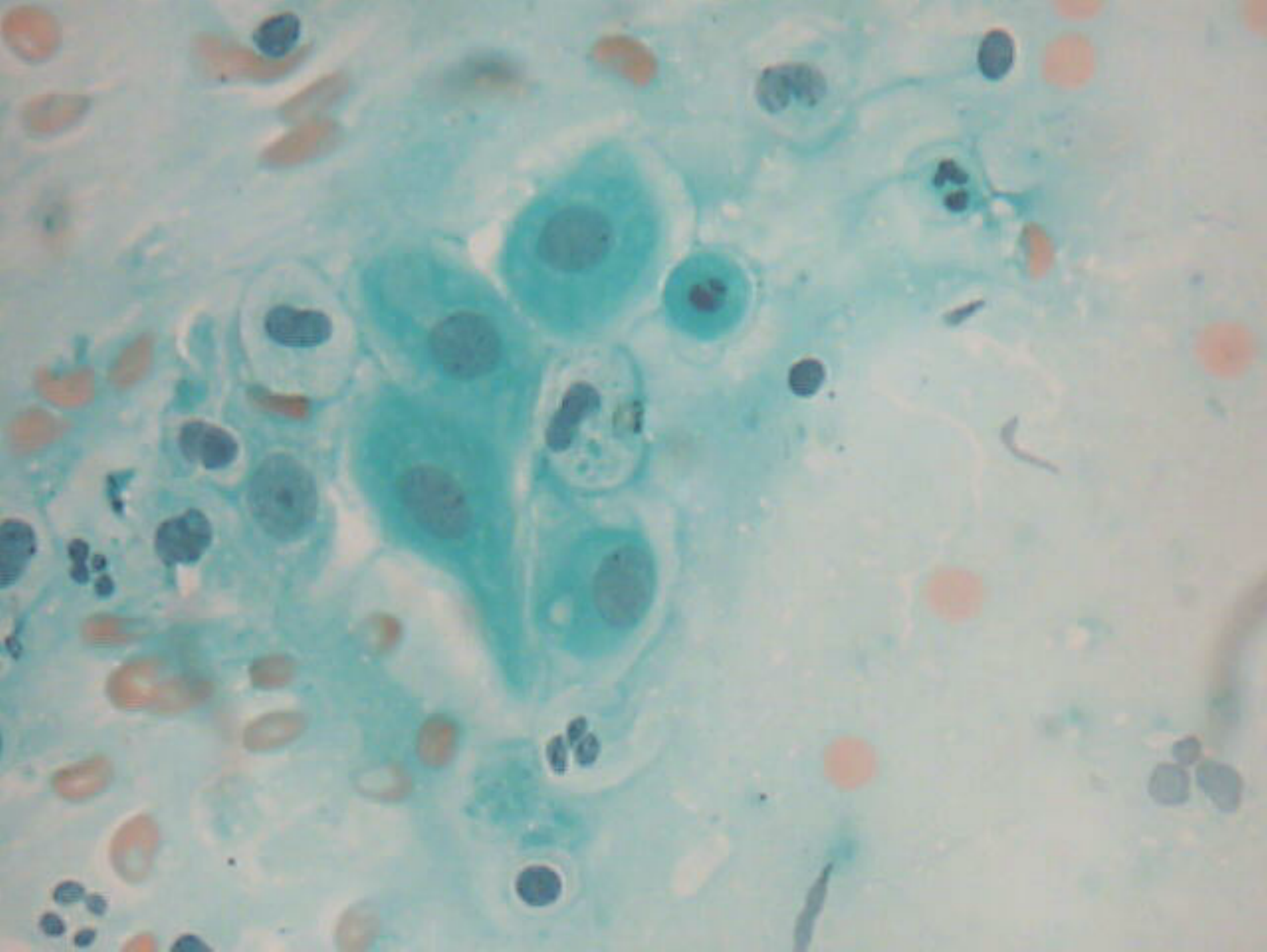}} \hfil
		\subfloat[]{\includegraphics[scale=0.13]{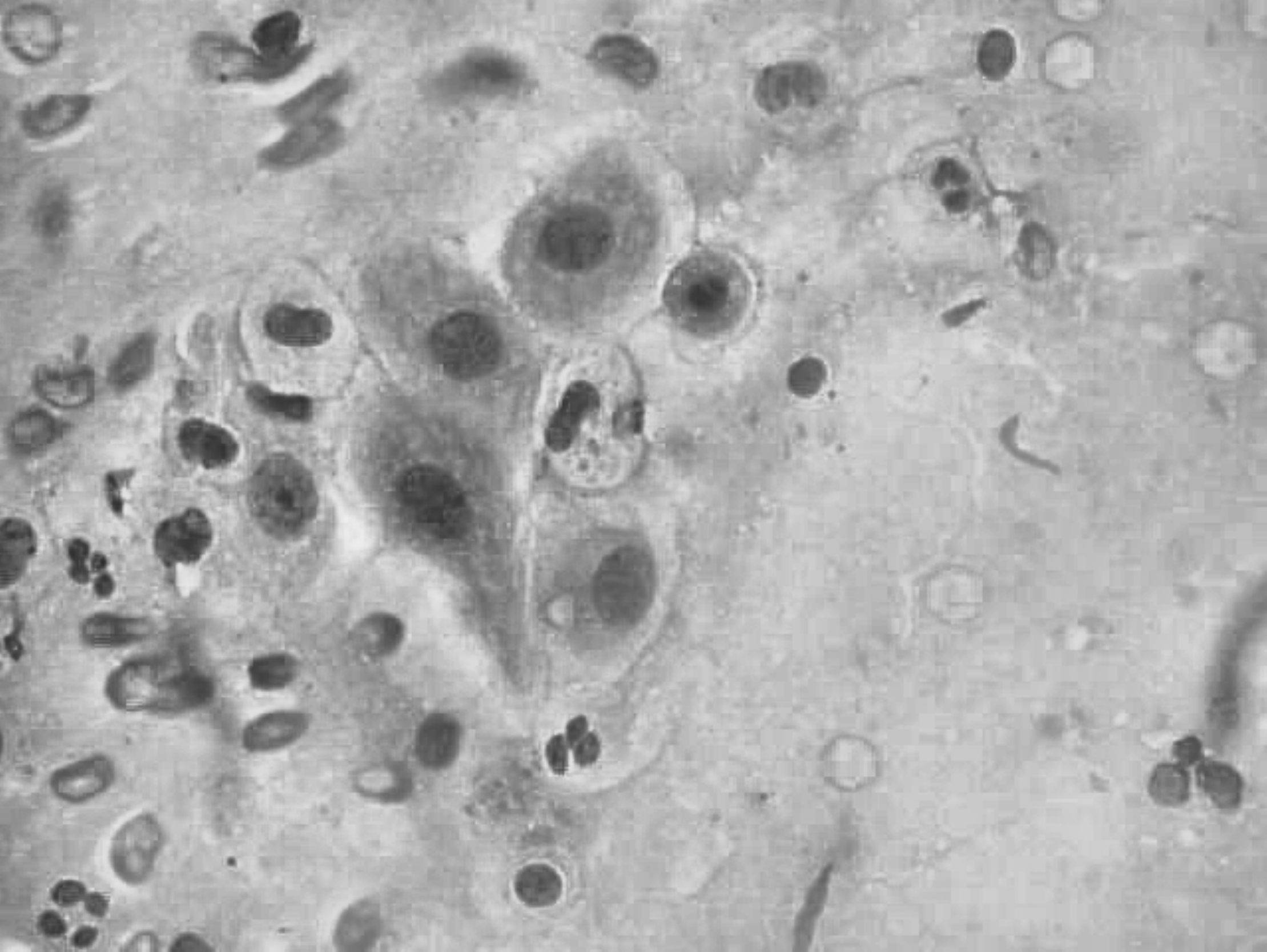}} \hfil
		\subfloat[]{\includegraphics[scale=0.13]{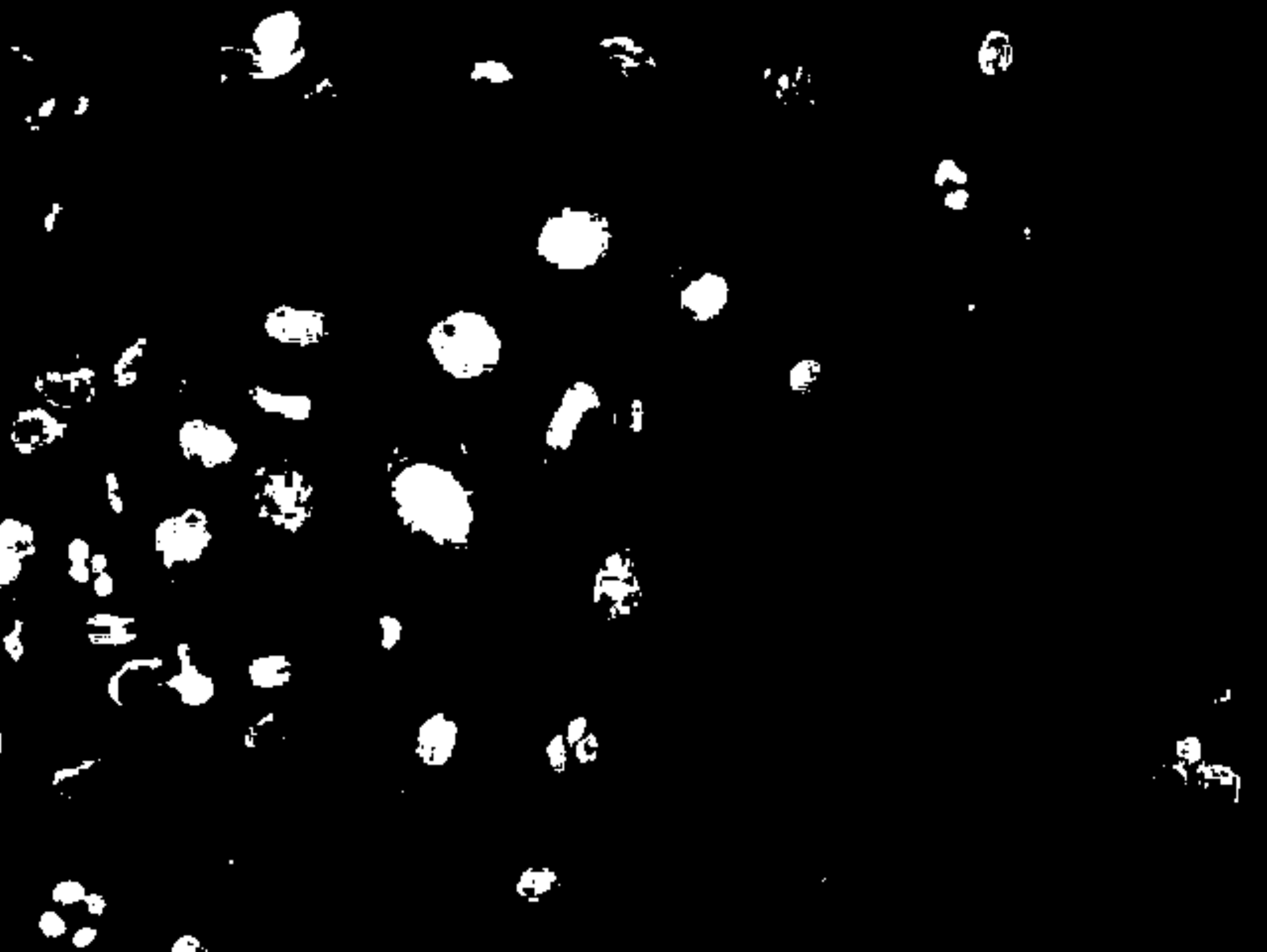}} \hfil
		\caption{\textbf{Nuclei detection} (a) Multi-cell image, (c) V channel after contrast adjustment by CLAHE, (d) Detected nuclei by a global threshold }
		\label{fig:detection}
		\vspace{-0.3cm}
	\end{center}
\end{figure}

For detection of nuclei in multi-cell images, we propose a straightforward algorithm applied on the V channel of HSV (Hue, Saturation, Value) color space of the images. Here we neglect the color information because of the presence of extreme color variation in the stains, as can be seen in Table \ref{tbl:Aindra}. The process is divided into three steps: 1) The PAP-smear images 
are generally noisy as can be seen in Figure \ref{fig:detection}. Hence, 
we apply median filtering with a $5\times5$ window. 
2) 
To improve the contrast and accentuate the differences between nucleus and background, we apply contrast-limited adaptive histogram equalization (CLAHE \cite{clahe}). 
3) Finally a global threshold, is applied which localizes the nuclei. 
We note that the detection is required only for the Aindra dataset. The Herlev dataset \cite{herlev}, consists of such detected single cell images.

\subsection{Segmentation of nuclei in single-cell PAP-smear images}
We propose a segmentation method comprising broadly of two steps i.e selective pre-processing followed by patch-based classification using CNN, as described in the following subsections. We have reported this method in \cite{myspie}. However, for self sufficiency, we also briefly discuss it here. We note that the approach assumes that the cell detection is already carried out, and thus operates on single-cell images. 
\subsubsection{Cell separation and selective pre-processing}
\label{ssec:sp}
\begin{figure}[!h]
	\begin{center}
		\subfloat[]{\includegraphics[scale=0.45]{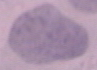}} 
		\subfloat[]{\includegraphics[scale=0.125]{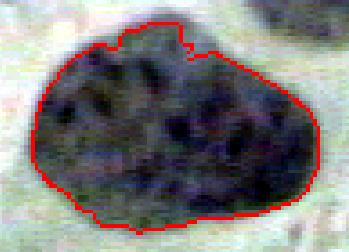}} \quad  
		\subfloat[]{\includegraphics[scale=0.1]{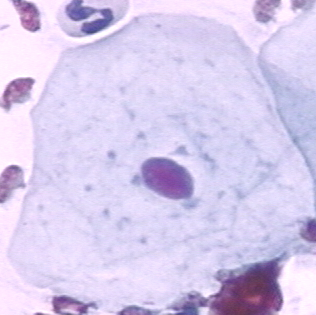}} 
		\subfloat[]{\includegraphics[scale=0.1]{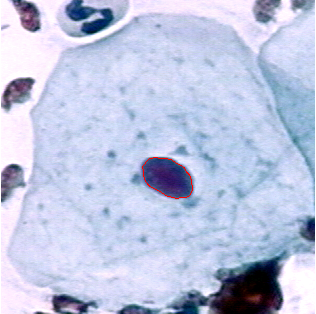}} 
		\caption{{
				\label{fig:pp} \textbf{Pre-processing}. (a, c) Original Image: Carcinoma in-situ, Normal intermediate (b, d) Respective pre-processed image with ground truth overlapped, 
			Note that pre-processing on (a) increases the contrast within the nuclei, while on (c) increases the contrast between the nuclei and the background.}}
	\end{center}
	\vspace{-0.5cm}
\end{figure}

Often, a contrast enhancement pre-processing aids the segmentation task. 
While this is useful for small and uniform nuclei, in this case, in most cells with larger nuclei due to the irregularity of chromatin pattern, pre-processing hinders good segmentation as it also increases the intra-nuclear contrast (within the nucleus) 
(see Figure ~\ref{fig:pp}). Thus, we suggest that the cells with small and compact nuclei need pre-processing, while those with bigger nuclei and irregular chromatin pattern don't. 

We propose feature-based cell separation method where we compute homogeneity \cite{glcm} of the original images, and use a threshold on its value for separating the cell images in two categories. 
Thus, images 
with relatively homogeneous and compact nuclei are passed on for pre-processing, before computing the final segmentation via CNN. For images with 
irregular nuclei, no pre-processing is done. 

\subsubsection{Patch-based CNN}
\label{ssec:cnn_seg}
For segmentation, we train two independent CNNs from scratch, one operating on patches from the pre-processed images and the other on patches from non-preprocessed images. During the testing phase, after cell separation, the images are passed on to the respective CNN. 
 We convert our 2-class classification problem (nucleus vs background) into a 3-class problem among nucleus (interior), edge (boundary) and background 
classes \cite{deepcell}. The details of the overall approach can be found in \cite{myspie}.

\subsection{Classification with detected nuclei}
In this section, we explore different strategies and scenarios for the classification of nuclei in cervical cell images.

\subsubsection{Multi-class classification using transfer learning}
Considering the success of deep learning methodologies in recent years for the task of classification, we too explore their application in our work. The PAP-smear images generally have a large appearance variation in terms of both contrast and color in the normal and abnormal cells. Furthermore, in medical imaging applications, very few annotated images are available, in the range of hundreds, as opposed to the millions of natural images available for other applications \cite{alexnet}. To overcome the aforementioned difficulties, we make use of the concept of transfer learning where the filters learned by a CNN, pre-trained on ImageNet consisting of millions of images, are directly used for classification in some other domain (medical images in our case). This strategy helps us in two ways: 
1)  It mitigates the dependency of deep CNNs on huge amount of annotated training data.
2) It effectively reduces the training time required for training a CNN from scratch.

\begin{figure}[!ht]
	\begin{center}
		\includegraphics[scale=0.18]{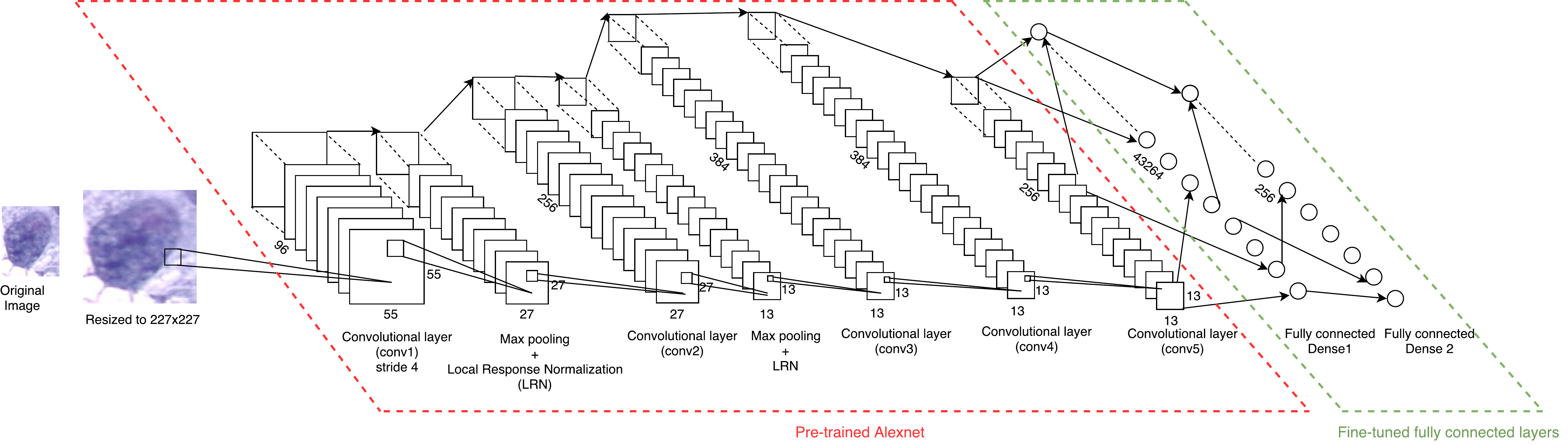}
		
		\caption{conv5T: features from fifth convolutional layer of Alexnet.}
		\label{fig:tlconv5}
	\end{center}
\end{figure}

\begin{figure}[!ht]
	\begin{center}
		\includegraphics[scale=0.22]{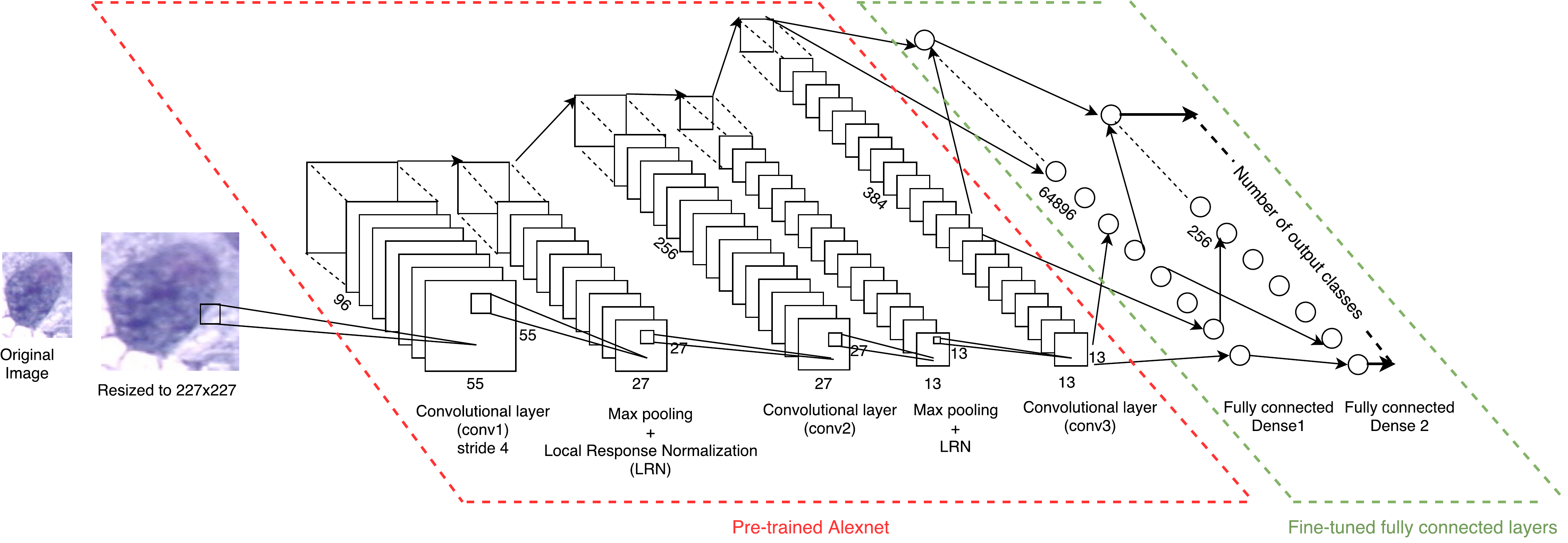}
		
		\caption{conv3T:features from third convolutional layer of Alexnet}
		\label{fig:tlconv3}
	\end{center}
\end{figure}

\begin{figure}[!h]
	\begin{center}
		\includegraphics[scale=0.3]{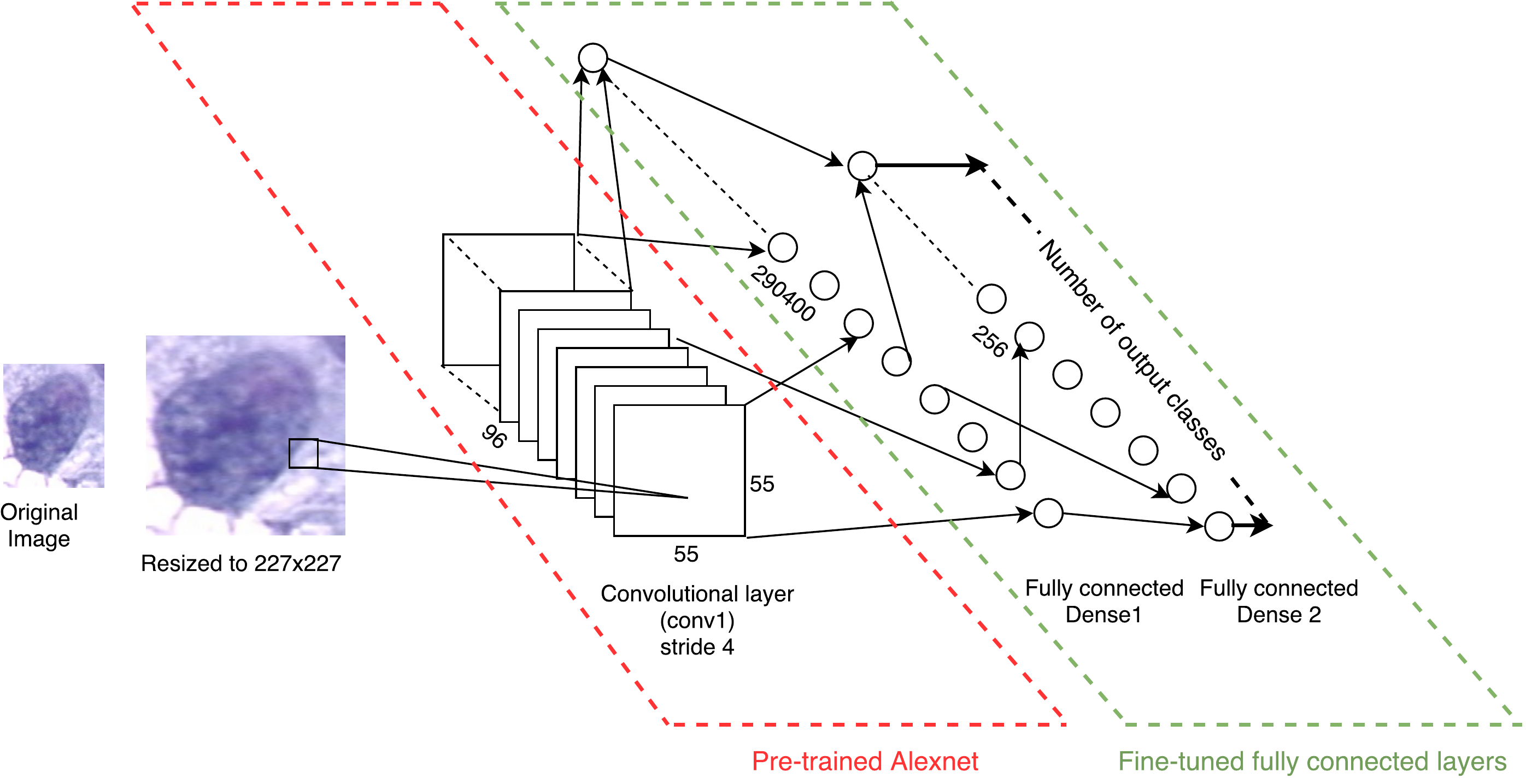}
		
		\caption{conv1T: features from first convolutional layer of Alexnet}
		\label{fig:tlconv1}
	\end{center}
	\vspace{-0.5cm}
\end{figure}

It is shown in literature that the lower level convolutional layers learn the low-level primitive features such as gradients, texture etc., and the deeper layers, learn the high-level data specific semantic features \cite{tl1}. Considering the hypothesis that semantic features may not be important for cell classification, we explore for classification, the outputs from the filters learned by Alexnet \cite{alexnet} at last (conv5), intermediate (conv3) and first (conv1) convolutional layers followed by two fully connected layers which we retrain, one consisting of 256 neurons and the last layer consisting of number of neurons equal to number of classes. We refer to these new transfer-learning based networks in the rest of the paper as conv5T (Figure \ref{fig:tlconv5}), conv3T (Figure \ref{fig:tlconv3}) and conv1T (Figure \ref{fig:tlconv1}).

\subsubsection{Decision-tree based classification using transfer learning}
\begin{figure}[!h]
	{
		\centering
		\includegraphics[scale=0.45]{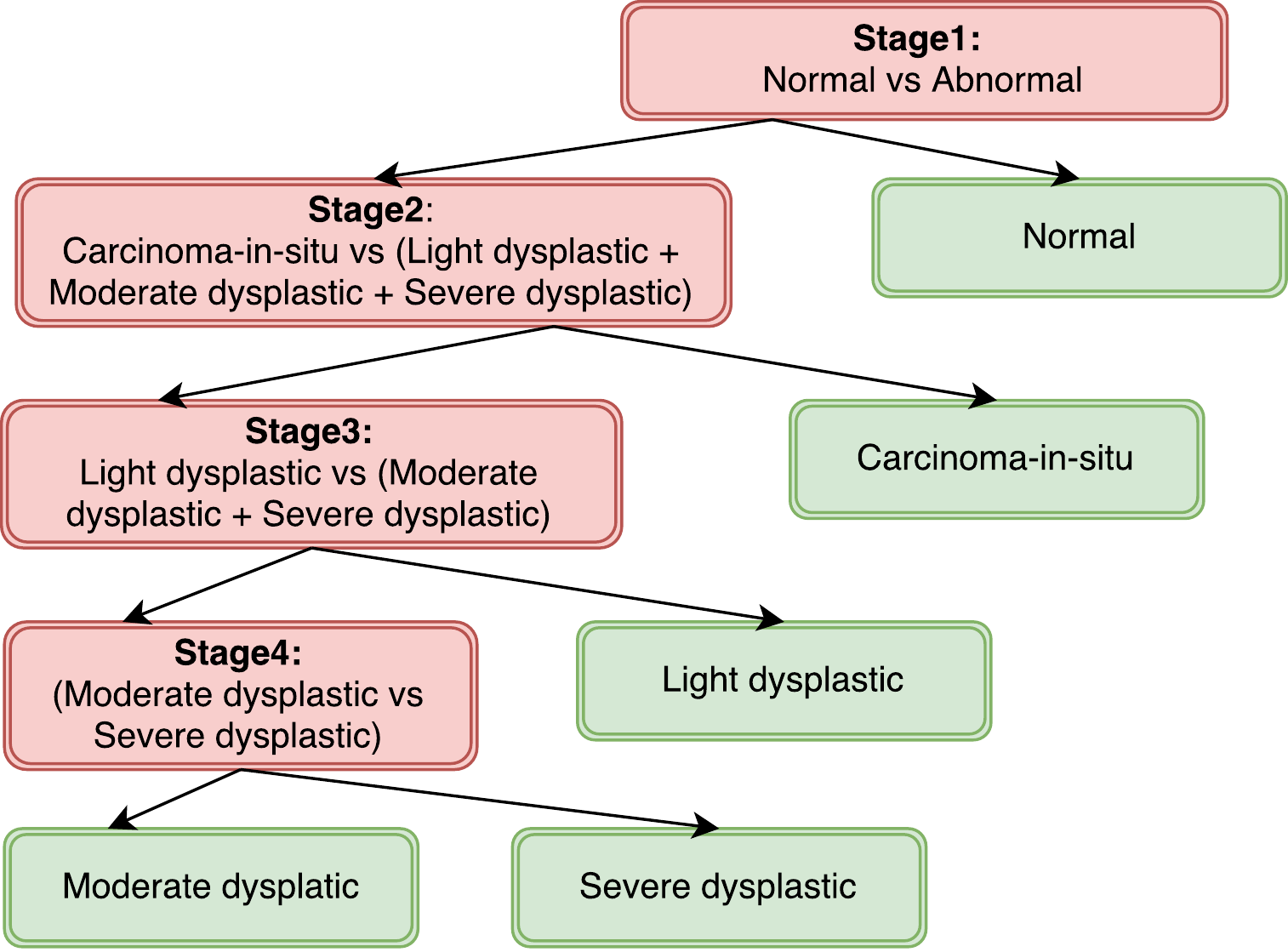}
		\caption{Decision-tree based approach for classification}
		\label{fig:decision_tree}
	}
\end{figure}

Considering certain similarities and differences between some classes, we propose a decision-tree based approach for classifying the cell images in a hierarchical way as shown in Figure \ref{fig:decision_tree}. 
At the first node, a two-class classification is done between the normal and abnormal classes. This is also important from the perspective of a screening system where only the difference between normal and abnormal classes can also be considered important. Additionally, once we have a good classification between the normal and abnormal class, we can classify the abnormal cells into further gradations of abnormalities. We achieve this in the daughter nodes where at the second level we discriminate between the highest level of abnormal class with other classes. Next, we discriminate between the lowest level of abnormal class with the remaining classes. Finally, at the leaf node, we discriminate between the leftover abnormal classes. The number of levels in the tree is based on the number of gradations of abnormalities in the dataset. At each node, we use a CNN with conv1T architecture (Figure \ref{fig:tlconv1}) for classification.

%

\subsubsection{Classification of detected nuclei in multi-cell images using transfer learning}
Following the success of CNN-based classification approach, we apply the transfer learning based methodology to the real-world multi-cell PAP-smear images. After detecting the nuclei with the help of detection algorithm mentioned, we extract the detected nuclei regions as sub-images using bounding boxes around the connected components. These sub-images are now passed on for classification to a CNN whose features for the first convolutional layer are extracted directly from the first convolutional layer of the pre-trained Alexnet. Because there can be large variations in multi-cell images, overfitting can occur. To reduce the chances of overfitting, we use two techniques: 1) By appending a few untrained convolutional layers before the fully connected layers, the number of nodes in the fully connected layers are reduced and hence the number of parameters to be trained in the network are reduced. 2) Using max-pooling dropout after every untrained convolutional layer \cite{maxpool_drop}.

\subsubsection{Classification with segmented nuclei}
For classification on segmented nuclei images with conv1T architecture. From the detected cells, the nuclei is segmented and the background values are replaced by 255. These images are now fed into the conv1T architecture, assisting it by emphasizing the nucleus features. 
Here, we also pose the question if the exact segmentation is needed for classification and demonstrate the answer through experimentation with and without segmentation of nuclei in single-cell images. 

\section{Experiments and Results}
We now discuss various experiments for detection, segmentation and different classification scenarios. In all the experiments involving supervised learning (segmentation and classification), we have used 70\% of images from each class are used for training, 15\% for validation and remaining 15\% for testing. The results are reported over a 5 random training, validation and testing sets.
\label{sec:exp}

\subsection{Evaluation metrics}
For segmentation, we have quantified the boundary of segmented nuclei using pixel-based F-score \cite{thresh1}.
%
%
For comparisons with other segmentation techniques, we use Zijdenbos similarity index (ZSI) \cite{MRF}.


For classification, based on the ground-truth label information in the datasets, we consider 2 problems i.e 2-class classification in Herlev and Aindra datasets and 7-class classification in Herlev dataset. We use accuracy
for quantification of classification approaches on both 2-class and 7-class classification problems. 
The overall accuracy is computed as the fraction of correctly classified cells over all classes \cite{thresh1}.

%

\subsection{Cervical cell nuclei detection results on Aindra dataset}
After obtaining the output from nuclei detection algorithm on multi-cell images, a bounding box with a padding of 20 pixels on 4 sides around each connected component is used to capture a sub-image containing the nuclei. These sub-images are labeled as normal/abnormal nuclei based on the ground truth annotations. The visual results for nuclei detection on Aindra dataset are shown in Table \ref{tbl:nd}. 
We observe that in most cases the detected cells indeed focuses on a single cell-nuclei, with some cell and background regions.
\begin{table}[!h]
	\centering
	\caption{Nuclei detection on Aindra dataset. Normal and abnormal detected nuclei are marked with green and red respectively.}
	\label{tbl:nd}
	\begin{tabular}{>{\centering\arraybackslash}p{5cm}|>{\centering\arraybackslash}p{1.2cm} >{\centering\arraybackslash}p{1.2cm}}
		\hline
		{\textbf{\scriptsize{Original Image}}} & \multicolumn{2}{c}{\textbf{\scriptsize{Detected nuclei}}} \\ \hline 
		\multirow{4}{5cm}{\begin{minipage}{5cm}	\centering\includegraphics[scale=0.30]{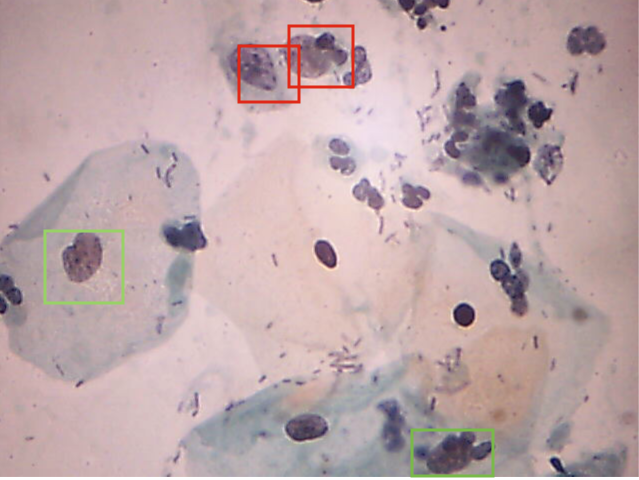} \end{minipage}} & \multicolumn{2}{c}{\scriptsize{Normal nuclei}} \\ 
		
		&
		\begin{minipage}{1.2cm}
			\centering\includegraphics[scale=0.45]{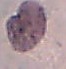}
		\end{minipage}
		& 
		\begin{minipage}{1.2cm}
			\centering\includegraphics[scale=0.45]{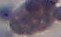}
		\end{minipage} \\
		
		& \multicolumn{2}{c}{\scriptsize{Abnormal nuclei}} \\ 
		& \begin{minipage}{1.2cm}
			\centering\includegraphics[scale=0.45]{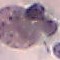}
		\end{minipage}
		& 
		\begin{minipage}{1.2cm}
			\centering\includegraphics[scale=0.45]{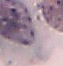}
		\end{minipage} \\
		\hline
		
	\end{tabular}
\end{table}

\subsection{Cervical cell nuclei segmentation results}
We provide the quantitative effectiveness of the proposed segmentation algorithm on Herlev dataset. We have used an architecture similar to that of VGGNet\cite{vggnet} for both our CNNs i.e CNN trained without pre-processed images (CNN$_{w1}$) 
and CNN trained with pre-processed images. (CNN$_{p1}$).


In Table \ref{tbl:classwise}, we compare final results of our approach (row1) with its counterpart without selective pre-processing i.e with 1) same pre-processing on all images (row3) 2) no pre-processing on any image (row2), which further stresses on the importance of cell separation step. We compare the results of CNN with 2 classes (nucleus and background, without the boundary class) with similar architecture as CNN$_{w1}$ and CNN$_{p1}$ and homogeneity-based cell separation technique (row 4). The results clearly show that a CNN with 2 class does not perform well. Therefore, the third class has significant effect in disambiguating the boundary class. 
We note that our method performs better than the state of the art FCM based technique \cite{11}, which also reports F-scores for individual classes. 

\begin{figure}[!h]
	\begin{center}
		\subfloat[]{\includegraphics[scale=0.25]{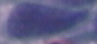}} 
		\subfloat[]{\includegraphics[scale=0.25]{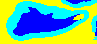}} 
		\subfloat[]{\includegraphics[scale=0.25]{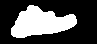}}
		\subfloat[]{\includegraphics[scale=0.25]{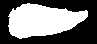}} \\
	
		\subfloat[]{\includegraphics[scale=0.1]{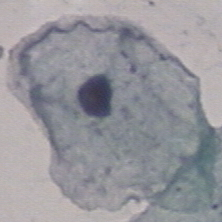}} 
		\subfloat[]{\includegraphics[scale=0.1]{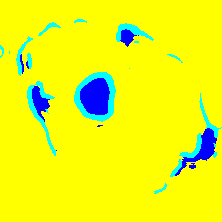}}  
		\subfloat[]{\includegraphics[scale=0.1]{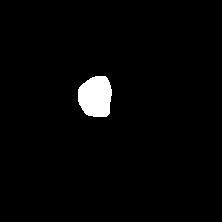}}
		\subfloat[]{\includegraphics[scale=0.1]{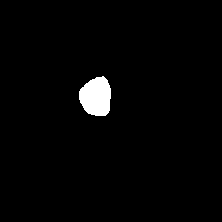}} \\
		
			\subfloat[]{\includegraphics[scale=0.20]{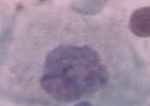}} 
		\subfloat[]{\includegraphics[scale=0.20]{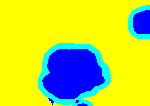}} 
		\subfloat[]{\includegraphics[scale=0.20]{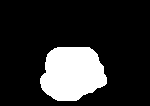}}
		\subfloat[]{\includegraphics[scale=0.20]{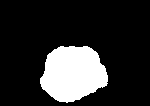}} \\
		
		\caption{\label{fig:final}
			\textbf{Nucleus segmentation}. 
			{
				(a),(e),(i) Original images of different sizes, (b),(f),(j) Segmentation map produced by CNN, where blue, cyan \& yellow color represents nucleus, edge and boundary pixels respectively, (c),(g),(k) Final segmentation map, (d),(h),(l) Ground truth. }}
	\end{center}
	\vspace{-0.5cm}
\end{figure}

We provide ZSI comparisons\cite{MRF} in Table \ref{tbl:zsi} with various contemporary methods which are also compared 
in \cite{MRF}. We note that the performance of the proposed approach is better than some of the contemporary methods (FCM\cite{11}, Threshold\cite{MRF} \& Hierarchical tree\cite{MRF}) and comparable to the best (MRF\cite{MRF} \& RGVF\cite{MRF}). 
Some visual results for the algorithm are shown in Figure \ref{fig:final}, where the segmentation map obtained is also shown with 3 classes i.e nuclei, edge and background. 

\begin{table*}[!htbp]
	\caption{Class-wise F-scores for nucleus segmentation results for Herlev dataset.}
	\label{tbl:classwise}
	\scriptsize
	\begin{center}
		\begin{tabular}{>{\arraybackslash}l| >{\centering\arraybackslash}c  >{\centering\arraybackslash}p{1cm} >{\centering\arraybackslash}c  >{\centering\arraybackslash}c >{\centering\arraybackslash}c>{\centering\arraybackslash}c >{\centering\arraybackslash}c >{\centering\arraybackslash}c}
			\hline
			
			 & {\scriptsize nsup}  & {\scriptsize nint} & {\scriptsize ncol}  & {\scriptsize ldys} & {\scriptsize mdys} & {\scriptsize sdys} & {\scriptsize cis} & {\scriptsize Average} \\
			\hline \hline
			

			
			
			

			{\scriptsize Proposed: feature-based CS with Homogeneity} 
			 & 0.86 & 0.91 & 0.89 & 0.92 & 0.94 & 0.87 & 0.89 & \textbf{0.90}\\ \hline

			
			{\scriptsize Without cell separation: no pre-processing on any image} 
			& 0.62 & 0.75 & 0.73 & 0.77 & 0.91 & 0.91 & 0.74 & \textbf{0.77} \\ \hline
			
			{\scriptsize Without cell separation: same pre-processing on all images} 
			 & 0.86 & 0.91 & 0.89 & 0.92 & 0.82 & 0.45 & 0.89 & \textbf{0.82} \\ \hline

			{\scriptsize 2-class CNN with homogeneity based CS} 
			 & 0.60 & 0.29 & 0.62 & 0.42 & 0.51 & 0.84 & 0.61 & \textbf{0.63} \\ \hline
			

			{\scriptsize FCM\cite{11}} 
			 & 0.84 & 0.89 & 0.83 & 0.83 & 0.84 & 0.83 & 0.78 & \textbf{0.84} \\
			
			
			\hline
			
		\end{tabular}
		
	\end{center}
\end{table*}

\begin{table*}[!t]
	\caption{Pixel-based ZSI comparison}
	\label{tbl:zsi}
	\scriptsize
	\centering
	\begin{center}
		\begin{tabular}{| c |  c |  c |  c | c| c |c|}
			\hline
			\textbf{\scriptsize Method} & \centering{\scriptsize Proposed} & {\scriptsize FCM \cite{11}}  & {\scriptsize Threshold \cite{MRF}} &  {\scriptsize Hierarchical tree \cite{MRF}} & {\scriptsize MRF \cite{MRF}} & {\scriptsize RGVF \cite{MRF}} \\ 
			
			\hline
			
			\scriptsize \textbf{ZSI} & \centering 0.90 & \centering 0.80 & \centering 0.78 & \centering 0.89 & \centering 0.93 & 0.93\\ \hline 
			
		\end{tabular}
		
	\end{center}
\end{table*}

\subsection{Cervical cell classification results}
\subsubsection{With detected nuclei using transfer learning}
\begin{figure*}[!ht]
	\begin{center}
		\includegraphics[scale=0.32]{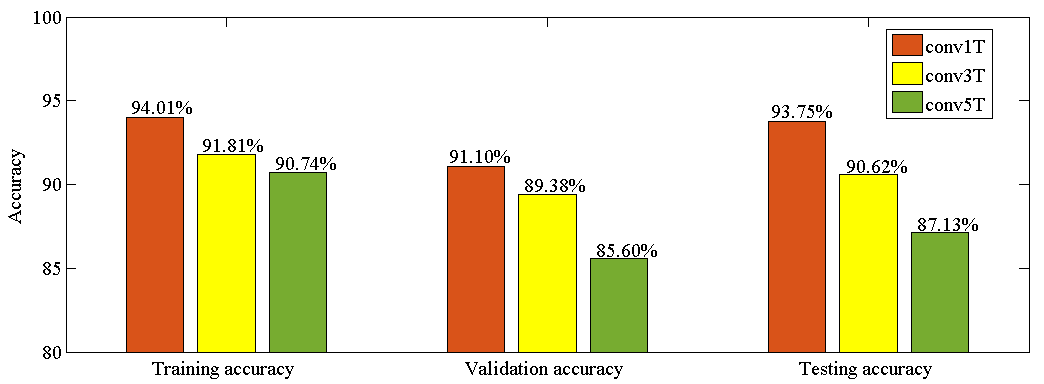}
		\caption{7-class CNN accuracies with transfer learning on Herlev dataset}
		\label{fig:tl1}
	\end{center}
\end{figure*}

For multi-class classification using transfer learning, we explore the architectures conv5T, conv3T and conv1T given in Figure \ref{fig:tlconv5}, \ref{fig:tlconv3} and \ref{fig:tlconv1} where we use the outputs from the fifth, third and first convolutional layers of Alexnet, respectively. After getting the respective outputs from  pre-trained Alexnet, we train the fully connected layers with a 256-neuron hidden layer and a final output layer with number of neurons equal to the number of classes. 
Because of high dimensional outputs from the convolution layers of Alexnet, the number of weights to be trained are huge (in the range of 1 million), hence we use data augmentation on the training data. We also use data augmentation on validation data to reduce the extreme fluctuations in validation accuracy while training. After this, we end up with 12,000 examples for training and 3000 for validation. We use 5 random sets of training, validation and testing data for the experimentation and report the average results in Figure \ref{fig:tl1}. All three of these networks are trained for 200 epochs 
and mean squared error as loss function. 

\begin{figure*}[!t]
	\begin{center}
	\subfloat{\includegraphics[scale=0.1]{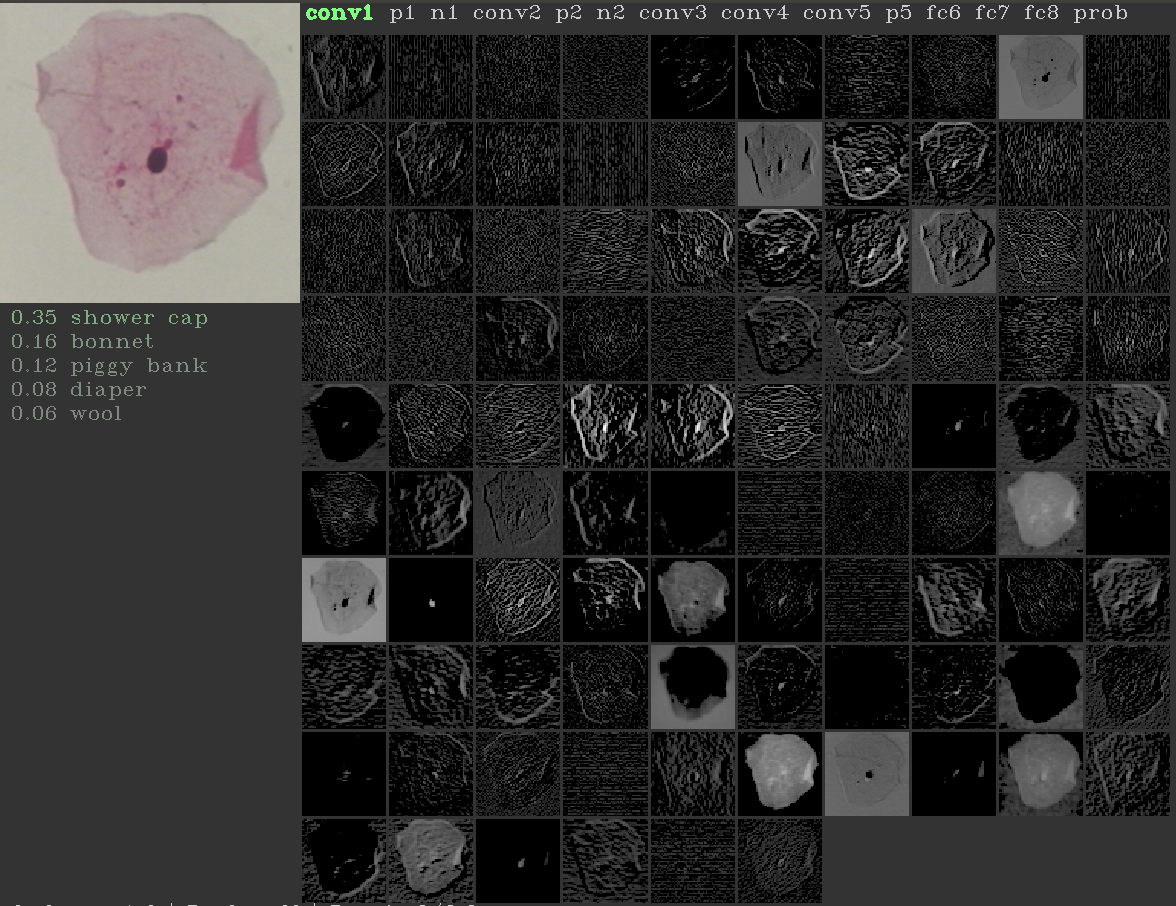}} \hspace{0.1cm}
		\subfloat{\includegraphics[scale=0.1]{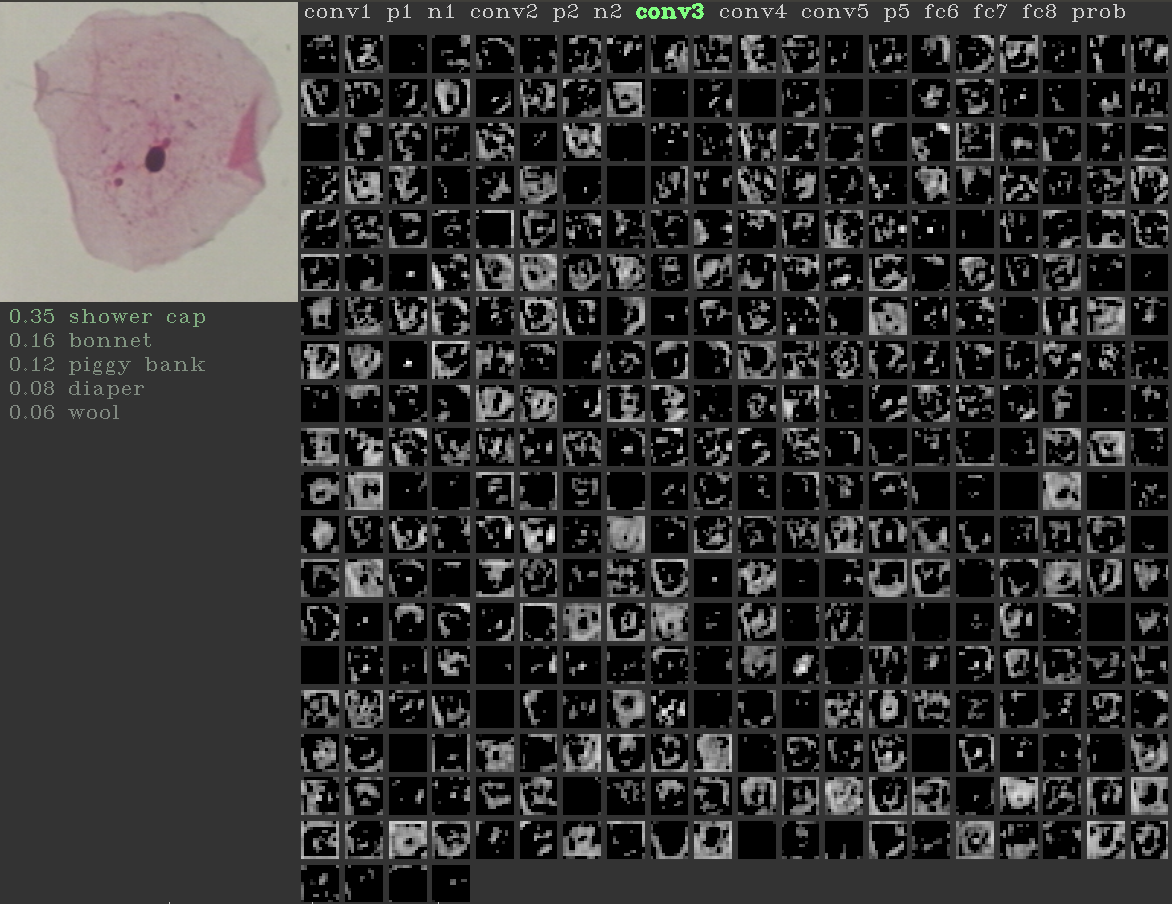}} \hspace{0.1cm}
		\subfloat{\includegraphics[scale=0.1]{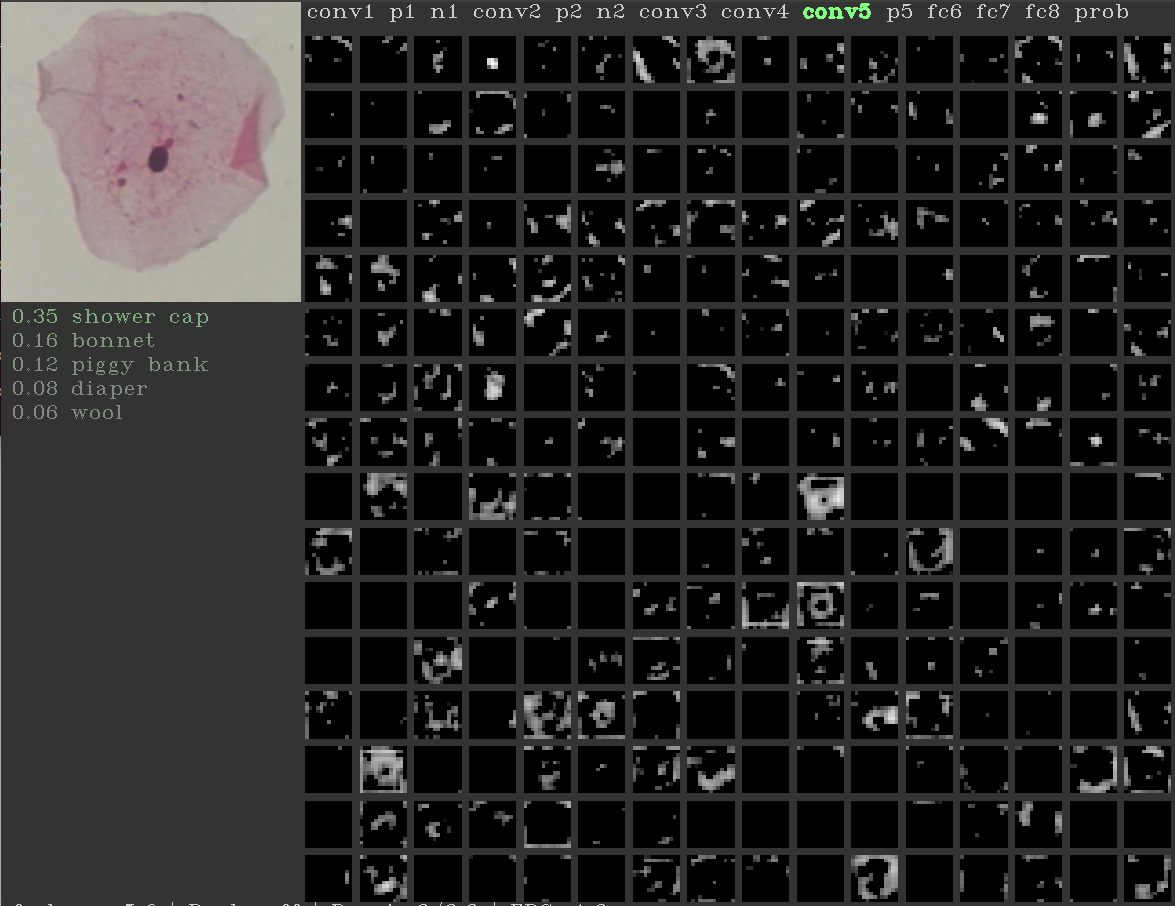}}
		\vspace{-0.3cm}
		\caption{\textbf{Activation maps:} Left to Right: Conv1, Conv3 and Conv5 activation maps for Normal superficial cell images.}
		\label{fig:am}
	\end{center}
\end{figure*}

Activation map from different convolutional layers of Alexnet for an example image are shown in Figure \ref{fig:am}. It can be seen that the activation map from the first convolutional layer learns the prominant texture features from the images as opposed to the third and fifth convolutional layers. 
This observation supports the hypothesis that for cell images, as the depth of the network increases, the high-level features do not seem informative. 
This also supports our motivation to select Alexnet consisting of smaller number of layers. We provide the average training, validation and testing accuracies for the 7-class classification, over 5 random trials for different architectures in Figure \ref{fig:tl1}. The constant increase in accuracies from conv5T to conv1T shows that the cell classification problem performs better with low-level features rather than those at the deeper levels. 
We believe this is an interesting and important insight, as typical deep learning approaches only consider the last layer features for classification.

\subsubsection{With detected nuclei using decision-tree}
Next, we explore the results of decision-tree based classification using transfer learning. Because of the transfer learning with conv1T giving the best results, we report the decision-tree based classification results with the architecture given in Figure \ref{fig:tlconv1}. 
The overall accuracies at each stage with transfer learning are given in Table \ref{tbl:tl2}. We note that the decision-tree based method with transfer learning gives high accuracy at each stage.
\begin{table}[!h]
	\centering
	\caption{Accuracies at different stage of decision-tree based approach using traditional and deep learning based methods.}
	\label{tbl:tl2}
	\scriptsize
	\begin{tabular}{|l|c|c|c|c|}
		\hline 
		& Stage 1 & Stage 2 & Stage 3 & Stage 4 \\ \hline 
		conv1T			        & 99.3\%    & 95.6\%     & 95.1\%  & 94.1\%   \\ \hline   
	\end{tabular}
\end{table}

\subsubsection{Classification on Aindra dataset}
After detection of nuclei in multi-cell images of Aindra dataset, we use transfer learning for classification. The connected components obtained from the nuclei detection step are extracted by taking a bounding box around the detected nuclei. A padding of 20 pixels from all four sides is applied on the actual seed region bounding box. These sub-images are labeled as normal/abnormal based on the overlap with the ground truth. Next, 3 random training and validation sets are created based on the whole slide images, which are passed on to the transfer learning architecture 
for classification into normal/abnormal classes. 
The mean training and validation accuracies are reported in Table \ref{tbl:tl_Aindra}. The training and validation accuracies are quite good considering that the Aindra dataset is complex in terms of contrast variation and artifacts.

\begin{table}[h]
	\centering
	\caption{Training and validation accuracies on Aindra dataset after nuclei detection}
	\label{tbl:tl_Aindra}
	\scriptsize
	\begin{tabular}{|l|c|c|c|}
		\hline 
		& Training accuracy & Validation accuracy   \\ \hline 
		Accuracy  & 95.5\%    &  85.7\%       \\ \hline   
	\end{tabular}
\end{table}

\subsubsection{With segmented nuclei}
Considering the best results with conv1T architecture, we now explore the test accuracies using conv1T with segmented nuclei Herlev dataset for 7 class classification. For this, we keep the original nuclei intensity values and set the background values as 255. 
The results of classification with ground-truth and proposed segmentations, and using the full cell images (without segmentation) are shown in Table \ref{tbl:conv1T_gt}. The results clearly demonstrate that with segmentation the performance is rather limited, and suggests that the contextual information in images with nuclei detections, may be important 
for good classification. Thus, it indicates that, as far as the classification performance is concerned, an easier nuclei detection process may be sufficient rather than a sophisticated segmentation approach. 

\begin{table}[!h]
	\centering
	\caption{Accuracies for segmented and non segmented cell images.}
	\label{tbl:conv1T_gt}
	\scriptsize
	\begin{tabular}{|p{1.5cm}|p{2cm}|p{2.5cm}|}
		\hline 
		Without segmentation & With ground truth segmentation & With segmentation using proposed method \\ \hline    
		93.75\%    & 82.33\%   & 78.25\%    \\ \hline   
	\end{tabular}
\end{table}

\subsubsection{Comparisons}
In Table \ref{tbl:comp}, we compare the results of our classification approaches on Herlev dataset for 2-class classification problem with the following existing methods: 1) Benchmark results in \cite{benchmark}, 2) Ensemble learning for three classifiers \cite{bora}, 3) Particle swarm feature selection and 1-nearest neighbor as classifier \cite{trad2}, 4) Genetic algorithm feature selection \cite{trad1}, 5) Artificial neural network with 9 cell-based features \cite{thresh1} and 6) Transfer learning by training a new architecture (ConvNet-T) from scratch \cite{deeppap}. We note that our results with conv1T surpass all of the existing algorithms and are quite close to the results with ANN in \cite{thresh1}. This might suggest that this is the best accuracy that can be reached for this dataset. For 7-class classification problem in Herlev dataset, we provide a comparison in Table \ref{tbl:comp2}. The results show that our approach surpasses the benchmark results and the results are extremely close to that of ANN \cite{thresh1}. 

We stress on the fact that our approach is comparatively easier (segmentation-free) and faster than that of \cite{thresh1} wherein both nuclei and cytoplasm are segmented, 
whereas we pass on the cell images directly to a CNN trained with transfer learning. Also, we note that our approach is arguably better than the similar approaches stated in \cite{deeppap} in terms of training time where they train their CNN architecture from scratch and we only train the fully connected layers.

\begin{table*}[!t]
	\caption{2-class classification (Normal vs Abnormal) on Herlev dataset}
	\label{tbl:comp}
	\vspace{-0.5cm}
	\centering
	\begin{center}
		\begin{tabular}{ | c|  c |  c |  c | c | c | c | c | }
			\hline
			 \textbf{\scriptsize Method} & {\scriptsize Proposed:conv1T} & {\scriptsize Benchmark \cite{benchmark}} &{\scriptsize Ensemble \cite{bora}} & {\scriptsize PSO-1nn \cite{trad2} } & \scriptsize{GEN-1nn \cite{trad1}} & \scriptsize{ANN \cite{thresh1}} &\scriptsize{ConvNet-T\cite{deeppap}}\\ 
			
			\hline
			
		  	\scriptsize \textbf{Accuracy} & \scriptsize\textbf{99.3}\% &  \scriptsize93.6\%  &  \scriptsize96.5\% & \scriptsize96.7\%& \scriptsize96.8\% & \scriptsize99.27\%  & \scriptsize98.3\%\\ \hline 
			
		\end{tabular}
	\end{center}
\end{table*}


\begin{table}[!t]
	\caption{7-class classification on Herlev dataset}
	
	\vspace{-0.5cm}
	\label{tbl:comp2}
	\begin{center}
		\begin{tabular}{| c |  c |  c |  c|}
			\hline
			\textbf{\scriptsize Method} & {\scriptsize Proposed (conv1T)} & {\scriptsize Benchmark \cite{benchmark}} & {\scriptsize ANN \cite{thresh1}} \\ 
			
			\hline
			
			\scriptsize \textbf{Accuracy} & \scriptsize93.75\% &  \scriptsize61.1\% & \scriptsize\textbf{93.78}\% \\ \hline 
			
		\end{tabular}
	\end{center}
\end{table}

\section{Conclusion}
\label{sec:conc}
In this paper, we reported a PAP-smear image analysis system for cervical cancer screening for both single and multi-cell images. The image analysis generally consists of three steps: detection, segmentation and classification. We propose a simple nuclei detection algorithm for multi-cell images, 
and a patch-based CNN approach with selective pre-processing for segmentation. This approach results in an overall F-score of 0.90 on Herlev dataset. For classification, we propose feature-level analysis using transfer learning on Alexnet on both single and multi-cell images. A decision-tree based classification is proposed as an alternative to the multi-class classification. Further, we prove through experimentation that accurate segmentation is not necessary for classification with deep learning. We obtain state-of-the-art classification accuracy on Herlev for 2-class (99.3\%) and for 7-class classification (93.75\%). 
\section*{Acknowledgment}
We acknowledge the support of Aindra Systems Pvt. Ltd. for funding this research and regular discussions.

\ifCLASSOPTIONcaptionsoff
  \newpage
\fi



\bibliographystyle{IEEEtran}
\bibliography{ref}
%
%
%

%







\end{document}